\documentclass[sigconf]{acmart}
\AtBeginDocument{%
  }

\setcopyright{acmlicensed}
\copyrightyear{2018}
\acmYear{2018}
\acmDOI{XXXXXXX.XXXXXXX}
\acmConference[Conference acronym 'XX]{Make sure to enter the correct
  conference title from your rights confirmation email}{June 03--05,
  2018}{Woodstock, NY}
\acmISBN{978-1-4503-XXXX-X/2018/06}

\usepackage{multirow}
\begin{document}

\title{SheetCompass: Hierarchical Relation Graphs for Agentic Spreadsheet Reasoning}
\author{
Panjing He, Mingyue Cheng, Yucong Luo,Li Li, Xiaohan Zhang
}
\affiliation{
  \institution{State Key Laboratory of Cognitive Intelligence, University of Science and Technology of China}
  \city{Hefei}
  \country{China}
}
\email{{hepanjing,lili0516,zxh25126485，prime666}@mail.ustc.edu.cn}
\email{mycheng@ustc.edu.cn}

\begin{abstract}

Spreadsheets are widely used to organize, analyze, and manipulate semi-structured data in real-world scenarios. However, automated spreadsheet reasoning remains a long-standing challenge for large language models (LLMs). Practical workbooks often feature a complex web of implicit cross-table associations and fine-grained column dependencies. To process them, existing methods typically flatten these multi-dimensional structures into sequential strings, which strips away the critical intra-sheet boundaries and inter-sheet semantics. As a result, LLMs are deprived of the global spatial layouts that human experts naturally leverage during visual scanning. To bridge this gap, we propose SheetCompass, a graph-guided and memory-driven agentic framework for spreadsheet automation reasoning. The key idea is to transform spreadsheets from raw tabular data into structured evidence spaces that can guide agent planning, execution, and reflection. Specifically, SheetCompass constructs a hierarchical graph to explicitly formalize implicit layouts into stable anchors at both the table and column levels, thereby mapping structural topologies while encoding spatial relations. To further enhance agent reliability, SheetCompass introduces a dual-level memory mechanism, which includes an expert knowledge memory that provides tool-use and domain knowledge and a reasoning experience memory that is dynamically updated from previous execution trajectories to support workflow evolution and prevent continuous mistakes. Based on these components, SheetCompass coordinates three specialized agents—the navigational explorer, logical programmer, and critical reflector—to handle perception, execution, and iterative verification. Driven by this multi-agent coordination, extensive empirical evaluations across various datasets demonstrate that SheetCompass achieves state-of-the-art performance in complex spreadsheet automation tasks. Our code is available at \url{https://anonymous.4open.science/r/sheetcompass-411C}.

\end{abstract}
\keywords{Natural Language Processing, Spreadsheet Manipulation, Multi-Agent Workflow}

\maketitle
\section{Introduction}
\begin{figure}[t]
    \centering
    \includegraphics[width=1\columnwidth]{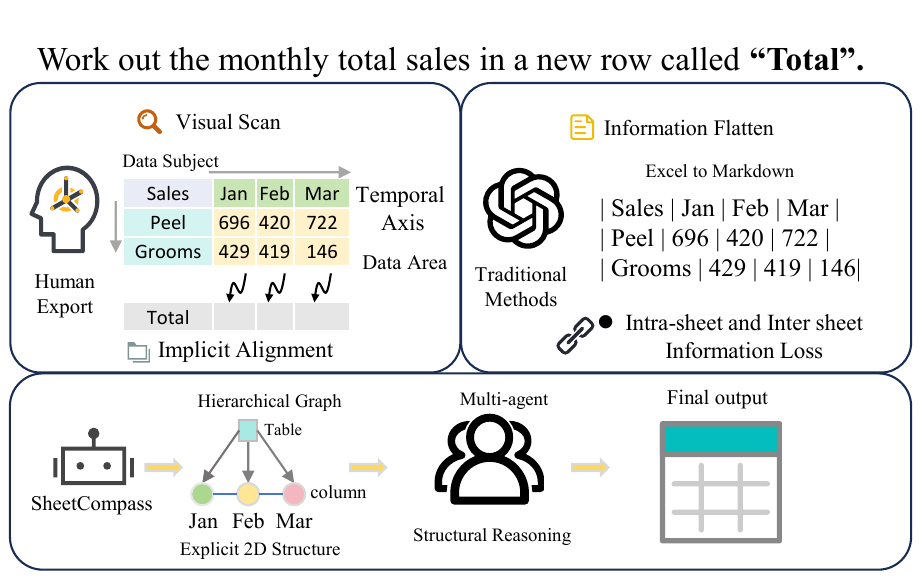}
    \caption{A comparison of various methods. Traditional markdown flattening collapses the 2D layout that human visual scanning effortlessly recovers. SheetCompass closes this gap with the hierarchical graph and multi-agent workflows.}
    \label{fig:intro}
\end{figure}

As a key knowledge representation medium, spreadsheets inherently include business logic and domain expertise\cite{intro1,intro2,intro3,intro5}. Nevertheless, the volume of heterogeneous data and the demand for repetitive manipulation make spreadsheet automation a critical bottleneck to increase industrial productivity\cite{re1,re2,m3}. In this context, Large Language Models (LLMs) emerge as a robust technical pathway for achieving automated spreadsheet\cite{intro4,re6,intro2.6}.

Although LLMs demonstrate superior semantic understanding, their performance suffers a sharp decline when processing complex spreadsheets\cite{intro2.1,inro2.2,inro2.3}. This weakness primarily stems from existing methodologies predominantly rely on an information-flattening paradigm, which linearizes multi-dimensional spreadsheets into sequential string representations such as Markdown or JSON. While preserving textual content\cite{intro2.5,intro2.6,m1,m2}, this approach fundamentally leads to the loss of intra-sheet topological structure and inter-sheet semantic understanding\cite{inro2.7,intro2.9,intro2.10,intro2.11}. On the intra-sheet level, sequential string fundamentally shatters the inherent orthogonal positioning system of spreadsheet rows and columns. Consequently, complex layouts lose their spatial neighborhood within the token stream\cite{inro2.7,intro2.12,intro2.13}. This separation makes it difficult for the model to understand the original grid layout of the spreadsheet\cite{intro2.8,intro2.14}. At the inter-sheet level, this flattening process hides the implicit data dependencies and business logic across different worksheets\cite{intro2.10,intro2.11}. Critical elements, such as the direct correspondences between cost and price sheets, are broken down into isolated text fragments\cite{intro2.9,intro2.13,intro2.14}. Without explicit inter-sheet semantic dependencies, the model fails to reconstruct the dynamic data flow across the workbook, leading to structural misalignment especially in cross-table reasoning\cite{intro2.6,intro2.11,intro2.14,intro2.15}.

This dual-level information loss reveals a fundamental mismatch with the original design intent of the spreadsheeas visual grid-based tables instead of linear text. Spreadsheets are a semi-structured medium grounded in visual cognition, not a flat collection of data\cite{intro3.4,intro3.5}. Its layout is aligned with the human visual processing system\cite{intro3.8}. As shown in Figure~\ref{fig:intro}, human experts rely on rapid visual scanning to instantaneously reconstruct the spatial topology of a spreadsheet. By leveraging spatial cues alone, they immediately identify column hierarchies, data entities, and distinct data regions. This intuitive process establishes an implicit alignment that bridges the semantic gaps between distant cells\cite{intro3.6,intro3.7,intro3.9}. The intricate intra-table topologies and inter-table dependencies of complex spreadsheets lead to profound information loss when fed into LLMs. Human experts utilize multi-dimensional layout perception\cite{intro3.11,intro3.12}, to extract latent semantic information. In contrast, models must rely on statistical structural approximation within linear sequences. This neglect of physical topology and the failure to recognize inter-sheet relationships forces the model to attempt complex reasoning in a dimensionally unconstrained sequential space. Consequently, this spatial understanding gap primarily impedes high-fidelity spreadsheet automation.

To address those information loss, this paper proposes the SheetCompass framework. This framework restores lost structural dependencies and ensures high-fidelity processing of complex tasks through a multi-agent workflow. For structural modeling, SheetCompass recasts neighborhoods into a hierarchical graph. Hierarchical edges explicitly connect table-level and column-level nodes to preserve both intra-sheet spatial layouts and inter-sheet semantic structures. Building on this recovered topology, the framework introduces a dual-level memory mechanism, including expert knowledge memory and reasoning experience memory to provide stable tool-use knowledge and drive dynamic trajectory evolution. Driven by these structured spaces and memories, a multi-agent workflow coordinates a suite of specialized agents, where the navigational explorer handles topological sensing and target anchoring, the logical programmer generates structured code execution, and the critical reflector conducts closed-loop validation. Empirically, across various complex spreadsheet datasets, SheetCompass substantially outperforms existing methods in task success rate. Ultimately, this structural-and-reasoning decomposition built around graph topology resolves this major bottleneck by enabling downstream agents to navigate multi-dimensional grids without losing their intrinsic grid properties.

Our contributions are threefold:

\begin{itemize}
    \item We propose SheetCompass, a novel framework that bridges structural space perception with multi-agent reasoning, effectively overcoming the structural information loss of conventional flat sequences.
    \item We introduce a hierarchical contextual graph that preserves intra-sheet spatial layouts and inter-sheet semantic linkages, combined with a dual-level memory and a coordinated multi-agent workflow to guarantee high-fidelity task execution.
    \item Extensive experiments on diverse complex benchmarks demonstrate that SheetCompass achieves state-of-the-art performance, empirically confirming the critical impact of structural and semanticon cross-table reasoning.
\end{itemize}

\section{Related Work}

\subsection{Automated Spreadsheet Manipulation}
Early work on spreadsheet automation centers on program synthesis for repetitive tasks\cite{re2,re3,re8,re9}. Representative systems such as FlashFill\cite{re2} search a domain specific language (DSL) for programs that satisfy user-provided specifications\cite{re10,re11,re12}. While these methods are highly precise for deterministic tasks, they suffer from a severe curse of dimensionality because the DSL search space expands exponentially as constraints increase\cite{re21}. This makes traditional program synthesis techniques struggle when handling complex tasks involving multi-tier dependencies and cross-table references\cite{re12,re20}. To mitigate rigid symbolic pipelines, later work uses deep learning to infer implicit spreadsheet structure\cite{re4,re5,re6,re16,re17}. SpreadsheetCoder introduces a neural formula prediction model that encodes row and column level context. TableSense\cite{re13} applies convolutional neural networks to layout detection. And TUTA\cite{re14} adopts a tree-structured transformer to model hierarchical dependencies. These methods move the field beyond brittle pattern matching toward semantic modeling, but cross-table logical reasoning remains challenging. LLMs push automation further toward intent understanding and autonomous reasoning\cite{re22,re18,re15}. SpreadsheetLLM\cite{re23} utilizes inverted index translation to compress layout formats, while SheetAgent\cite{re18} and SheetCopilot\cite{intro3} use multi-agent pipelines that split tasks into subtasks and call a code interpreter. Most LLM-based pipelines still serialize sheets as flat text\cite{re19}. That representation hides nonlinear topology and makes it harder to match how human experts work.

\subsection{LLM-based Agents}

Early research predominantly focused on augmenting the logical reasoning capabilities of individual LLMs\cite{inro2.2,re2.2}. While prompting methods like Chain-of-Thought (CoT) enhance problem-solving by unlocking step-by-step reasoning \cite{re2.4,re2.5}, modern frameworks like ReAct \cite{re6} carry this multi-step paradigm further by anchoring reasoning in empirical action. This integration enables agents to dynamically interact with their environments, alternating between internal thought generation, external tool execution, and subsequent feedback adaptions \cite{re2.7,re2.8}. Nevertheless, when confronted with long-horizon tasks of extreme complexity, single-agent systems frequently encounter the error accumulation effect inherent in protracted reasoning chains\cite{re2.9}. Furthermore, they are susceptible to falling into self-verifying logical closures, which precipitates severe hallucinations and a fundamental lack of systemic robustness\cite{re2.10,re2.11}. To transcend the performance bottlenecks and reliability constraints of monolithic models, the research community has progressively pivoted toward Multi-Agent Systems (MAS)\cite{re2.12}. This paradigm reconfigures complex global objectives into orchestrated sub-task workflows by introducing structured role-playing and granular labor-division mechanisms such as MetaGPT\cite{re2.14} and ChatDev\cite{re2.15}. By leveraging inter-agent debate\cite{re2.16} or collaboration\cite{re2.18}, MAS can effectively rectify the inherent biases and stochastic errors of individual models\cite{re2.19}. This evolutionary shift from isolated individual intelligence toward synergetic collective intelligence provides profound theoretical inspirations for constructing more resilient, specialized, and interpretable research methodologies within highly dynamic and complex scenarios\cite{re2.20}.

\begin{figure*}[t]
    \centering
    \includegraphics[width=1\textwidth]{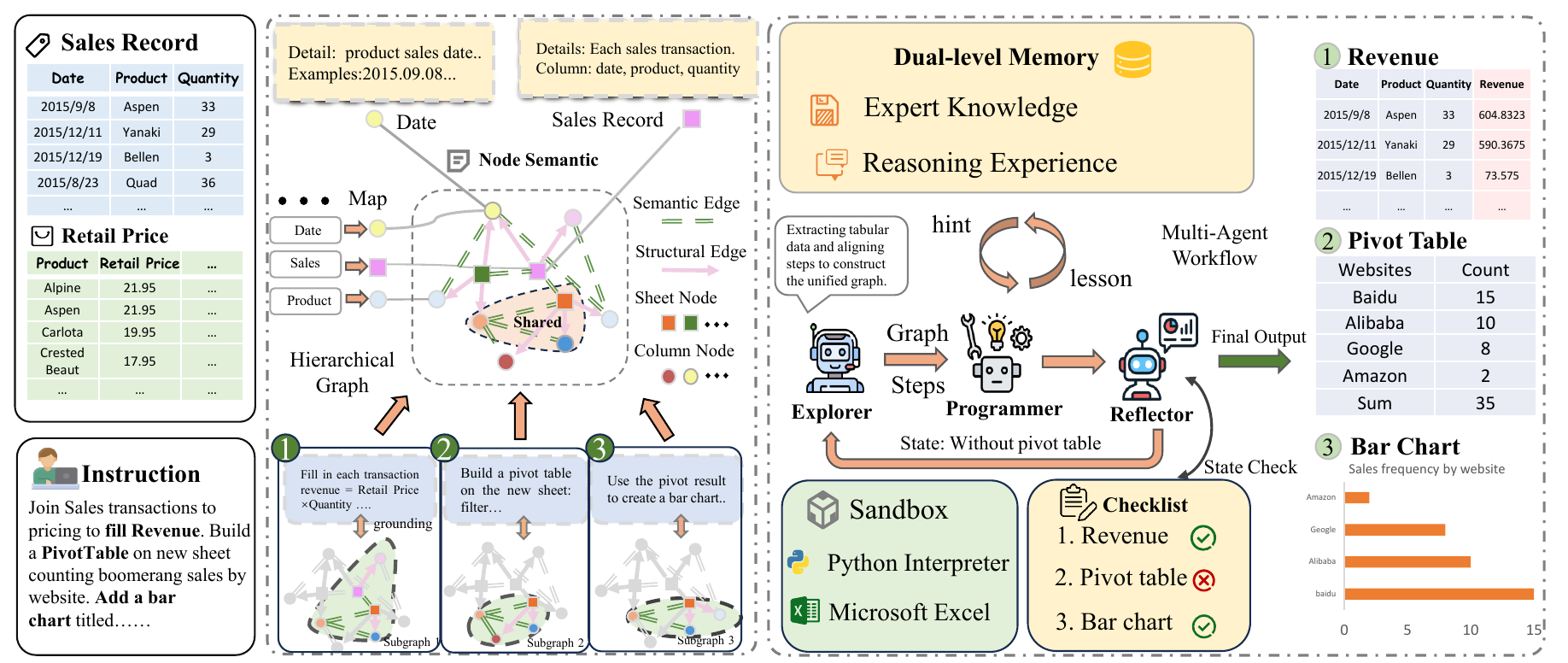}
    \Description{A flowchart showing the SheetCompass framework, starting from data input to the final alignment output.}
    \caption{An overview of the SheetCompass framework.}
    \label{methods}
\end{figure*}

\section{Methods}
In this section, we present the SheetCompass framework. We first formalize the problem of automated spreadsheet manipulation, followed by a detailed elaboration on our methodological design. Specifically, we introduce our coordinate-based structural perception through comprehensive graph representations, followed by the memory mechanisms and the subsequent logical reasoning workflow driven by our multi-agent.

\subsection{Problem Formulation}
Let $\mathcal{T} = \{T_1, T_2, \dots, T_M\}$ denote the initial set of spreadsheets, where $M$ denotes the total number of distinct tables across all sheets and each sheet may contain multiple tables. Each user-issued instruction $q$ specifies a series of automated operations to be performed on $\mathcal{T}$. Formally, spreadsheet automation is defined as $\mathcal{D} = \{(\mathcal{T}, q, \mathcal{A}^*)_{i}\}_{i=1}^{|\mathcal{D}|}$, where $\mathcal{A}^*$ is the ground-truth response consisting of the target tables. The goal of the task is to model the probability of a generated table sequence $\hat{\mathcal{A}}$ conditioned on the input collection $\mathcal{T}$ and the query $q$, formulated as:
\begin{equation}
    \hat{\mathcal{A}} = \arg \max_{\mathcal{A}'} p(\mathcal{A}' \mid q, \mathcal{T}; \theta)
\end{equation}
where $\theta$ denotes the parameters of the model.

\subsection{Overview of the SheetCompass Framework}
To bridge the gap in structured grid processing, the SheetCompass framework decouples spreadsheet automation into space reconstruction and coordinated reasoning. As shown in Figure\ref{methods}, the framework first rebuilds the spreadsheet's missing structural dependencies by transforming raw grids into a unified hierarchical graph. By mapping columns and cross-sheet relationships into connected topological nodes, this graph successfully preserves both localized layout geometry and global semantic connections. Guided by this reconstructed space, SheetCompass employs a dual-level memory system, combining static expert knowledge with dynamic reasoning experience, to provide the foundation for robust decision-making. This rich pool of structural and cognitive knowledge ultimately drives a collaborative multi-agent workflow. Within this execution loop, specialized agents dynamically navigate the graph topology to anchor targets, generate precise execution scripts, and perform closed-loop self-reflection. Through this integration of topological mapping, dual memory guidance, and multi-agent synergy, SheetCompass ensures high-fidelity execution across multi-dimensional spreadsheet tasks without losing the intrinsic properties of the data.

\subsection{Hierarchical Graph Construction}

\subsubsection{Graph Modeling}The fundamental challenge in automated spreadsheet reasoning lies in the severe loss of structural information during data serialization. Traditional approaches typically flatten multi-dimensional grids into text sequences, discarding critical spatial layouts and global dependencies across multiple tables. To resolve this, we model the spreadsheet as a hierarchical graph, formally defined as $\mathcal{G} = (\mathcal{V}, \mathcal{E})$. Here, the vertex set $\mathcal{V}$ encompasses multi-level spreadsheet components, while the edge set $\mathcal{E}$ maps out the intricate relationships among them. To mimic how human experts naturally comprehend data, we divide these edges into structural edges ($E_{\text{str}}$) for physical layouts and semantic edges ($E_{\text{sem}}$) for logical associations. This representation effectively transforms scattered cells into a connected knowledge space, providing global navigation and contextual constraints for downstream reasoning.

Guided by the human habit of visual scanning, we organize the vertex set $\mathcal{V}$ into a two-tier hierarchy consisting of table nodes and column nodes. Specifically, for each table node, we extract its constituent columns to establish ownership and structural dependencies. A column node $v_i \in V_{\text{col}}$ within the table is then defined by pairing its header name with a representative sample of its raw data entries. By avoiding dense cell-level modeling, this design successfully captures both semantic identity and factual context. Upon establishing these node representations, we automatically construct structural edges ($E_{\text{str}} \subseteq \mathcal{E}$) directly from the sheet layout using deterministic parsing rules. This edge set $E_{\text{str}}$ primarily consists of two functional categories: inclusion edges, which link each table node to its constituent column nodes based on structural boundaries, and adjacency edges, which connect neighboring column nodes according to their horizontal order, denoted as $\text{pos}(i)$ and $\text{pos}(i+1)$. Together, these rule-extracted edges form the stable backbone of our graph. 

\subsubsection{Semantic Alignment}The main upgrade in SheetCompass is how it connects related columns, even if they are far apart in the spreadsheet. To do this, we first give each column node a rich background by header name and actual data rows into a feature vector $\mathbf{x}$. Next, we use a pre-trained transformer $\Phi$ to map these features into representations $\mathbf{h}_i = \Phi(\mathbf{x}_i)$ and calculate their spatial-semantic similarity score using the cosine similarity function, denoted as $sim_{ij} = (\mathbf{h}_i \cdot \mathbf{h}_j) / (\|\mathbf{h}_i\| \|\mathbf{h}_j\|)$.To prevent mistakes, we also ask an LLM to score their logic relationship ($\mathcal{F}_{\text{LLM}}$), checking things like primary and foreign keys. We then mix these two scores together using a simple formula to make the final decision: \begin{equation} \epsilon_{ij} = \begin{cases} 1, & \text{if } \left[ \beta \cdot sim_{ij} + (1 - \beta) \cdot \mathcal{F}_{\text{LLM}}(v_i, v_j) \right] > \alpha \\ 0, & \text{otherwise.} \end{cases} \end{equation} Here $\alpha$ is the confidence threshold and $\beta$ decides which score matters more. If the combined total passes the threshold, we draw a logical link ($\epsilon_{ij} = 1$) between the columns. Otherwise, we ignore it. This clear cut-off filters out wrong guesses and leaves only high-quality connections. Ultimately, this turns the messy spreadsheet into a clean, smart network, helping our downstream agents quickly find the right data across different sheets.

\subsection{Dual-level Memory}
To support multi-step reasoning under iterative feedback, SheetCompass introduces the expert knowledge memory as a permanent, general knowledge base. This component equips the dynamic agent ensemble with both stable domain knowledge and reliable tool-use knowledge, transforming risky text generation into precise engineering execution. Specifically, domain knowledge organizes procedural skills into clear rules, such as frequent formulas and common charts. Meanwhile, tool-use knowledge extracts general repair lessons by logging historical code-execution histories, which systematically captures the root causes of errors alongside their correct programming solutions. By looking up these verified rules and templates, the framework establishes a solid cognitive baseline that prevents large language models from making wrong guesses during complex spreadsheet tasks.

In contrast, the reasoning experience memory is initially limited to the single task, functioning as a dynamic record of the active reasoning session. It carefully logs the episodic details of the current task, including where the agent moved on the graph, the error messages from the sandbox code, and the mismatches found by the checker. Feeding this short-term history directly back into the next prompting rounds ensures that each new action builds correctly upon previous outcomes, keeping data perception and automated reasoning tightly aligned. Crucially, the experiences accumulated during this active reasoning process are not discarded. Instead, successful trajectories and high-value reasoning lessons are extracted and transferred into long-term storage, continuously expanding the core repository of the expert knowledge memory.

\subsection{Multi-agent Workflow}
To transform complex instructions into executable code, SheetCompass coordinates a multi-agent workflow consisting of three dedicated roles: the explorer, the programmer, and the reflector. The explorer is designed to resolve the challenges of task decomposition and spatial locating within heterogeneous spreadsheet structures. It first maps high-level user instructions $q$ onto a sequence of logically dependent atomic steps $\mathcal{T} = \{t_1, t_2, \dots, t_n\}$. Upon determining the target worksheet set, the explorer extracts topological features from the hierarchical graph $\mathcal{G}$ that are highly correlated with each sub-task $t_i$. To achieve this, it constructs a seed node set $\mathcal{S}_i$ using a filtering mechanism based on semantic similarity:
\begin{equation}
\mathcal{S}_i = \{ v_j \in V \mid \cos(\mathbf{t}_i, \mathbf{h}_j) \geq \lambda \},
\end{equation}
where $\lambda$ represents the seeding coefficient that acts as the filtering threshold. This mechanism ensures that seed nodes possess high semantic confidence while maintaining a compact context size, effectively preventing prompt overflow. Subsequently, the system performs adaptive subgraph evolution originating from $\mathcal{S}_i$. By running a breadth-first search (BFS) on $\mathcal{G}$ and finding the intersection of subgraph sequences, the globally shared subgraph $\mathcal{G}_{\text{shared}} = \bigcap_{i=1}^n \mathcal{G}_i$ is extracted. This approach significantly compresses the context length.

The programmer acts as a constrained logic generation engine, whose primary responsibility is to combine the globally shared subgraph to transform each step $\tau_i$ into executable scripts. The Programmer follows a strict graph-grounded constraint mechanism. Unlike traditional text-to-code generation, the programmer must check entity alignment to ensure that all generated code variables are strictly anchored within the verified node sets of the spreadsheet graph. Within a secure sandbox environment equipped with a python interpreter and an excel engine, the programmer iterates through the tasks, producing explicit planning and action trajectories for each round. This restricts the search space of code generation to the verified physical layout of the spreadsheet, maintaining semantic consistency during multi-step reasoning.

The reflector implements a state-check feedback mechanism to ensure the robustness of task execution. Its core logic is to translate the explicit constraints within the user instruction $q$ into a concrete checklist, and then validate whether the metadata state $\mathcal{M}$ of the spreadsheet after running the code satisfies each item. Upon detecting any discrepancy between the actual execution state $\Delta(\mathcal{M})$ and the checklist requirements, the reflector transforms the logical contradictions into clear diagnostic notes and feeds them back to the workflow to trigger an iterative refinement process. Within this closed-loop feedback logic, a hyperparameter $\tau$ defines the maximum number of correction cycles allowed. Each time the reflector intercepts an error and requests a realignment from the explorer, one cycle is recorded, continuing until the task satisfies all items on the checklist or the threshold $\tau$ is reached.

\begin{table*}[ht]
  \centering
  \caption{Comparison of the performance of SheetCompass with other approaches on the SCB, SB, and SheetRM benchmarks(\%). $\uparrow$ indicates that a higher value is better. The best results are in bold and the second best are underlined.}
  \label{main_results}
  \begin{tabular}{llcccccc}
    \toprule
    \multirow{2}{*}{Backbone} & \multirow{2}{*}{Method} & \multicolumn{2}{c}{SCB} & \multicolumn{2}{c}{SB} & \multicolumn{2}{c}{SheetRM} \\
    \cmidrule(lr){3-4} \cmidrule(lr){5-6} \cmidrule(lr){7-8}
    & & Exec@1 $\uparrow$ & Pass@1 $\uparrow$ & Soft $\uparrow$ & Hard $\uparrow$ & Exec@1 $\uparrow$ & Pass@1 $\uparrow$ \\
    \midrule
    \multirow{6}{*}{GPT-4} 
    & Binder       & / & 15.6 & 1.2 & 0.0 & / & 6.3 \\
    & VBA          & 77.8 & 37.1 & 8.3 & 5.2 & 56.2 & 2.8 \\
    & OS-Copilot   & / & 50.1 & 17.0 & \underline{14.0} & 63.7 & 12.3 \\
    & SheetCopilot & 87.3 & 44.3 & 13.8 & 10.1 & 68.1 & 0.0 \\
    & SheetAgent   & \underline{94.1} & \underline{61.1} & \underline{17.6} & 13.5 & \underline{92.4} & \underline{31.2} \\
    & Ours         & \textbf{95.1} & \textbf{63.2} & \textbf{22.6} & \textbf{18.3} & \textbf{93.7} & \textbf{42.5} \\
    \midrule
    \multirow{6}{*}{GPT-5}   
    & Binder       & / & 60.0 & 18.1 & 14.7 & / & 8.1 \\
    & VBA          & 86.4 & 42.2 & 10.3 & 8.4 & 61.3 & 5.4 \\
    & OS-Copilot   & / & 60.0 & 18.0 & 14.0 & 74.8 & 20.5 \\
    & SheetCopilot & 65.0 & 55.0 & 16.5 & 13.1 & 52.7 & 2.2 \\
    & SheetAgent   & \underline{90.0} & \underline{70.0} & \underline{18.3} & \underline{15.1} & \underline{89.3} & \underline{44.8} \\
    & Ours         & \textbf{96.4} & \textbf{71.3} & \textbf{24.7} & \textbf{22.0} & \textbf{94.2} & \textbf{52.3} \\
    \bottomrule
  \end{tabular}
\end{table*}

\section{Experiments}
To validate the overall effectiveness of SheetCompass and the contributions of its core components, we conduct comprehensive experiments, including main comparisons against strong baselines and detailed ablation studies.

\subsection{Experimental Setup}

\subsubsection{Datasets}
To comprehensively evaluate the performance of the proposed framework, we adopted the SCB\cite{intro3},  SB\cite{intro2.6},and SheetRM\cite{re18} dataset as our experimental benchmarks. Within all datasets, each instance is characterized by a complex sheet structure. Furthermore, for the SB dataset, each task instance is accompanied by three independent test cases to rigorously assess the robustness and accuracy of the model's reasoning outcomes.
\subsubsection{Metrics}

To evaluate our framework comprehensively, we employ tailored metrics for the SCB and SB benchmarks. For SCB and SheetRM dataset, we utilize exec@1 to measure the runtime success rate and pass@1 to verify functional correctness against the ground truth\cite{intro3}. For SB dataset, which contains multiple independent test cases per task, we adopt soft restriction and hard restriction metrics. The soft restriction calculates the average success rate across all test cases, evaluating general competence without heavily penalizing rare edge cases. Conversely, the hard restriction requires a solution to perfectly pass all test cases to receive credit\cite{intro2.6}.

\subsubsection{Baselines}
To evaluate the effectiveness of SheetCompass, we select several representative baseline methods that cover two main technical paths: static generation and dynamic interaction. First, we include Binder, a classic table QA approach that translates natural language into SQL or Python code for data retrieval. We also incorporate traditional VBA script generation schemes, which use LLMs to generate macro code for native excel interfaces. These methods follow a static paradigm characterized by one-time generation. While they perform well on simple and fixed tasks, they cannot perceive or correct errors during execution. To evaluate more advanced interactive capabilities, we compare our framework against LLM-based agents, including SheetCopilot, SheetAgent, and OS-Copilot. These frameworks use a closed-loop architecture based on a plan-action-observation cycle, allowing them to break down complex instructions into steps and dynamically adjust their actions based on environment feedback.
\subsubsection{Backbones}
We choose two representative LLMs from the GPT family as the backbones to evaluate model performance on different scales. Specifically, we select GPT-5 as our primary closed-source LLM due to its exceptional reasoning and instruction-following capabilities in complex spreadsheet tasks. In addition, we employ GPT-4o-mini as a highly cost-efficient backbone. This choice allows us to verify the generalizability and robustness of SheetCompass under more constrained model capacities.
\begin{table*}[t]
  \centering
  \caption{Ablation study of the SheetCompass framework across different datasets (\%). The results quantify the performance contributions of the hierarchical graph, dual-level memory, and multi-agent workflow, coupled with a fine-grained analysis of their internal components. $\uparrow$ indicates that higher values represent better performance.}
  \label{ablation}

  \begin{tabular}{llcccccc}
    \toprule
    \multirow{2}{*}{\textbf{Category}} & \multirow{2}{*}{\textbf{Component}} & \multicolumn{2}{c}{SCB} & \multicolumn{2}{c}{SB} & \multicolumn{2}{c}{SheetRM} \\
    \cmidrule(lr){3-4} \cmidrule(lr){5-6} \cmidrule(lr){7-8}
    & & Exec@1 $\uparrow$ & Pass@1 $\uparrow$ & Soft $\uparrow$ & Hard $\uparrow$ & Exec@1 $\uparrow$ & Pass@1 $\uparrow$ \\
    \midrule
    Full Model & SheetCompass & \textbf{96.4} & \textbf{71.3} & \textbf{24.7} & \textbf{22.0} & \textbf{94.2} & \textbf{52.3} \\
    \midrule
    \multirow{3}{*}{Main Component} & w/o Hierarchical Graph & 87.2 & 56.4 & 18.1 & 14.5 & 85.2 & 41.4 \\
    & w/o Multi-agent & 89.5 & 59.1 & 19.3 & 15.2 & 87.4 & 43.3 \\
    & w/o Dual-level Memory & 92.6 & 60.4 & 21.2 & 18.8 & 90.5 & 44.3 \\
    \midrule
    \multirow{2}{*}{w/ Hierarchical Graph} & w/o Structural Edge & 90.8 & 62.3 & 20.5 & 16.8 & 88.7 & 45.7 \\
    & w/o Semantics Edge & 95.0 & 68.2 & 21.0 & 17.1 & 92.8 & 50.0 \\
    \midrule
    \multirow{2}{*}{w/ Dual-level Memory} & w/o Expert Knowledge & 93.8 & 69.7 & 22.7 & 17.3 & 91.7 & 51.1 \\
    & w/o Reasoning Experience & 94.5 & 68.9 & 23.1 & 18.4 & 92.3 & 50.5 \\
    \midrule
    \multirow{2}{*}{w/ Multi-agent} & w/o Explorer & 92.4 & 64.8 & 21.5 & 18.0 & 90.3 & 47.5 \\
    & w/o Reflector & 94.5 & 61.5 & 22.8 & 16.2 & 92.3 & 45.1 \\
    \bottomrule
  \end{tabular}
\end{table*}
\subsection{Experimental Results}
As presented in Table \ref{main_results}, our proposed SheetCompass consistently outperforms all baseline methods across all benchmarks, demonstrating substantial and robust improvements. Under the GPT-4 backbone, SheetCompass advances the state-of-the-art by improving pass@1 on SCB to 63.2\%,  hard restriction on SB to 18.3\% and pass@1 on SheetRM to 43.5\% ,which noticeably surpass the strongest baseline. This performance edge becomes even more pronounced when scaling to the GPT-5 backbone. Notably, on the challenging SB dataset, SheetCompass achieves absolute improvements of 6.4\% on soft restriction and 6.9\% on hard restriction.

This steady performance edge is primarily driven by SheetCompass's architecture, which effectively bridges structural perception with iterative execution. While existing baselines leverage LLM planning, they typically operate on flattened data sequences and struggle to capture the complex, non-linear topology of spreadsheets. In contrast, SheetCompass explicitly models the spreadsheet environment through its Hierarchical Graph, maintaining a high-fidelity structural skeleton while preserving cross-cell semantic dependencies. Guided by this topological grounding, the multi-agent workflow coordinates distinct roles to construct a holistic reasoning loop. Instead of relying on directly code generation, SheetCompass dynamically adapts to execution feedback while remaining anchored to stable structural constraints, thereby achieving state-of-the-art performance in spreadsheet automation.

\subsection{Ablation Study}

\subsubsection{Effect of Main Components}
The ablation results presented in Table \ref{ablation} indicate that each major component of the SheetCompass framework contributes significantly to the overall performance. Completely removing the hierarchical graph causes the sharpest performance drop, lowering the pass@1 on SCB and SheetRM by 14.9\% and 10.9\% respectively, while reducing the hard restriction on SB by 7.5\%. This substantial decrease demonstrates that the structure provided by the graph is a foundational source of grounding rather than an optional enhancement. Similarly, disabling the multi-agent workflow or the dual-level memory leads to noticeable performance degradation across all metrics. These findings confirm that high-fidelity structural perception, collaborative multi-agent execution, and continuous memory retention are all indispensable and complementary to the robustness of SheetCompass.

\subsubsection{Effect of Hierarchical Graph}
Further analysis of the hierarchical graph reveals the relative importance of its internal structural cues. Removing the structural edges reduces the SheetRM exec@1 from 94.2\% to 88.7\%, which suggests that the table-column topology effectively preserves essential layout information. This layout preservation allows the model to capture the coarse organization of spreadsheets without facing excessive token overhead from modeling every individual cell relation. On the other hand, removing semantic edges severely harms performance on the challenging SB hard restriction task, where the score drops from 22.0\% to 17.1\%. This indicates that topological structure alone is insufficient when complex tasks require tracking cross-column dependencies and implicit formulas.

\subsubsection{Effect of Dual-level Memory}
The dual-level memory provides essential task guidelines and interactive execution experiences necessary for long-term reasoning stability. Disabling expert knowledge reduces the SB hard restriction from 22.0\% to 17.3\%, which confirms that complex logical constraints in spreadsheets require explicitly injected professional rules to guide the LLM. Furthermore, removing reasoning experience causes the SCB pass@1 score to drop from 71.3\% to 68.9\%. This performance loss indicates that learning from historical execution trajectories helps SheetCompass avoid repetitive mistakes and handle long-range dependencies more effectively during multi-step tasks.

\subsubsection{Effect of Multi-agent Workflow}
At the execution stage, the multi-agent workflow coordinates distinct roles to generate and validate solutions through a collaborative process. As shown in Table \ref{ablation}, removing this workflow lowers the SCB pass@1 to 59.1\%, confirming that a single-path generation strategy is inadequate for complex operations. The fine-grained analysis reveals a clear division of labor between the agents. Removing the explorer decreases the SCB pass@1 to 64.8\%, as the system loses the heuristic exploration needed to discover plausible execution paths. Meanwhile, removing the reflector primarily harms performance on the SheetRM pass@1, driving it down to 45.1\%. This drop underscores the critical role of self-correction, as complex spreadsheet automation depends heavily on an independent agent to verify code execution against structural constraints.

\begin{figure}[t]
    \centering
    \includegraphics[width=\linewidth]{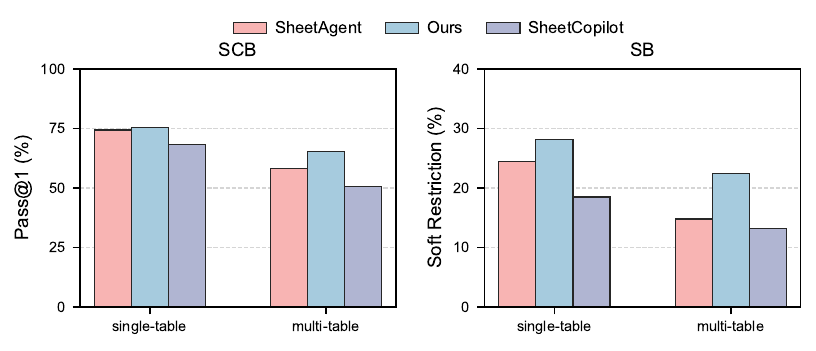}
    \caption{Performance comparison between single-table and multi-table scenarios across various baseline.}
    \label{single-table}
\end{figure}

\begin{figure*}[t]
    \centering
    \includegraphics[width=\textwidth]{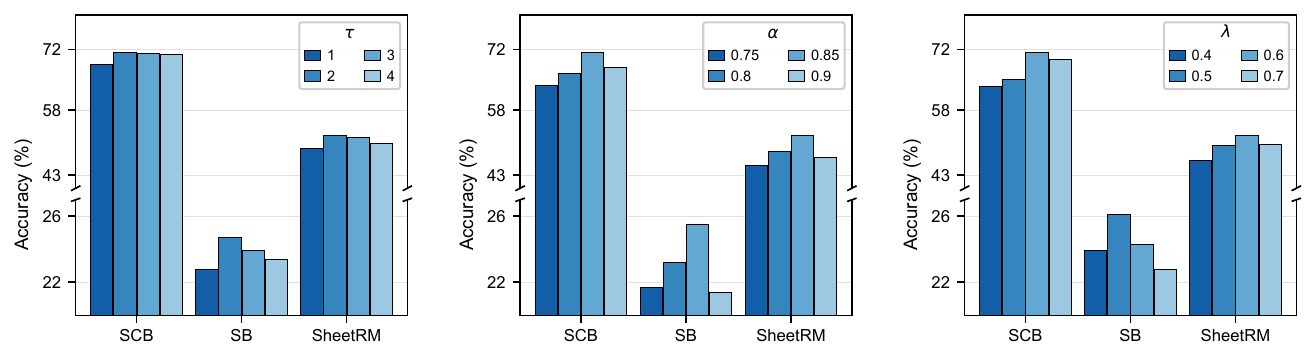}
    \caption{Hyperparameter sensitivity analysis for SheetCompass. The bars represent accuracy on SCB (left), SheetRM(middle), SB (right) datasets across different settings for reasoning cycles ($\tau$), confidence threshold ($\alpha$), and seeding coefficient ($\lambda$).}
    \label{sensitivity_analysis}
\end{figure*}

\subsection{Experimental Analysis}

\subsubsection{Analysis of Task Complexity}
To assess how SheetCompass handles complex spreadsheets, we evaluate its performance under both single-table and multi-table settings. As shown in Figure \ref{single-table}, all methods perform better in the single-table setting than in the multi-table setting, reflecting the added challenge of cross-table dependencies. On SCB, the pass@1 of SheetAgent drops from 74.4\% to 58.1\% when moving from a single table to multiple tables. This drop indicates that linear spreadsheet inputs make it difficult for traditional methods to track columns across different worksheets. In contrast, SheetCompass is less affected by this layout change, achieving the highest score of 65.2\% in the multi-table setting with a much smaller performance drop. A similar trend is observed on the SB dataset, where SheetCompass outperforms the baseline by 7.7\% under the soft restriction metric. This is because SheetCompass builds a hierarchical graph over tables and columns, providing explicit structural anchors that help the model locate target data and filter out irrelevant columns.

\begin{figure}
    \centering
    \includegraphics[width=\linewidth]{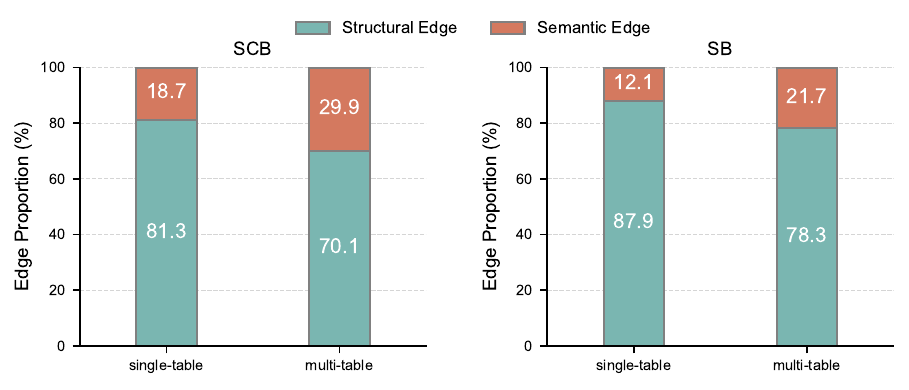}
    \caption{Distribution of structural and semantic edges across SCB and SB datesets.}
    \label{edge}   
\end{figure}

Figure \ref{edge} shows how the edge distribution of the hierarchical graph changes as task difficulty increases. In the single-table SCB setting, structural edges dominate at 81.3\%, indicating that the model primarily relies on the physical skeleton of the spreadsheet when tasks are local. This is expected, as most single-table operations can be resolved using nearby headers and basic table structures. However, in the multi-table setting, the proportion of semantic edges rises from 18.7\% to 29.9\%. The SB dataset shows a similar trend, with semantic edges increasing from 12.1\% to 21.7\%. These results demonstrate that physical layout alone is insufficient for complex cross-table tasks where data connections are hidden. SheetCompass handles this by maintaining these semantic edges, which also helps the multi-agent workflow operate more effectively. Specifically, the Explorer can narrow down the operation space, allowing the system to accurately translate these dependency paths into executable operations.

\subsubsection{Hyperparameter Sensitivity Analysis}
The hyperparameter $\tau$ controls the number of reasoning cycles performed by the multi-agent workflow in SheetCompass. Increasing $\tau$ from 1 to 2 improves the accuracy on SCB and SheetRM to 71.9\% and 54.7\% respectively. This trend indicates that a single inference pass is often insufficient for understanding complex table logic. Through repeated cycles of generation, feedback, and revision, the model can fix runtime failures and semantic errors found by the reflector. However, when $\tau$ increases to 3 or 4, performance plateaus or slightly declines. For most spreadsheet tasks, two cycles are enough for error correction. Further iterations increase computational costs and can introduce unhelpful code edits, trapping the model in local optima. Therefore, we set $\tau=2$ as the default value to balance accuracy and efficiency.

\begin{figure*}[t]
    \centering
    \includegraphics[width=1\textwidth]{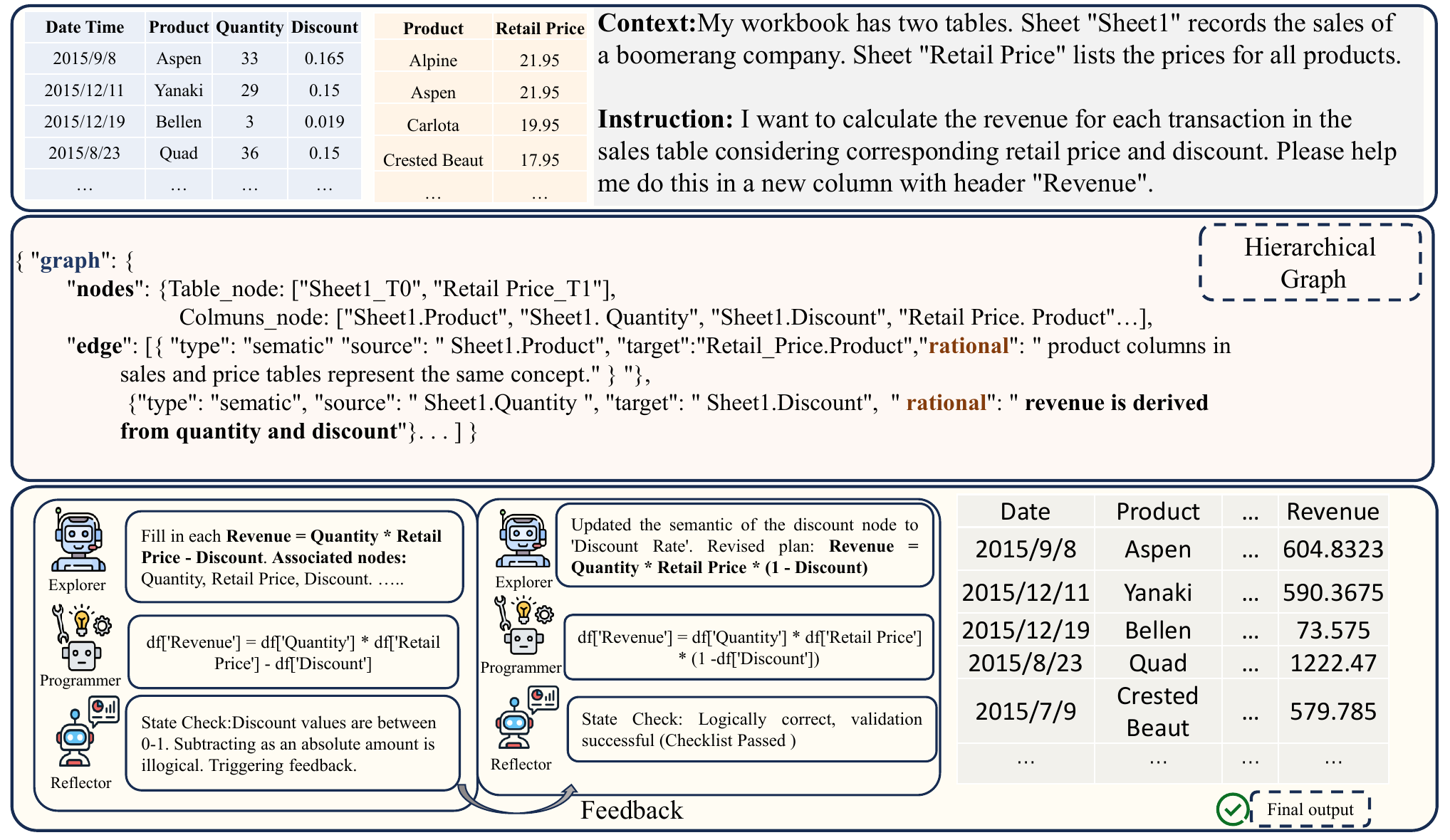}
    \caption{A case study of SheetCompass solving a cross-table task. The figure illustrates the transition from spreadsheet  to the construction of a hierarchical graph, followed by the iterative reasoning and verification process within the multi-agent.}
    \label{case}
\end{figure*}
The parameter \(\alpha\) serves as the confidence threshold for building the hierarchical graph. A candidate node is integrated into the topology only when its score passes this boundary. The framwork accuracy shows a concave pattern relative to $\alpha$, peaking at 0.85 with scores of 52.6\% on SheetRM and 25.5\% on SB. At a lower threshold ($\alpha=0.75$), the graph retains too many low-confidence nodes. These noisy nodes obscure critical structural lines and mislead the downstream reasoning process. Conversely, at a higher threshold ($\alpha=0.9$), many important structural vertices are excluded. This creates a sparse topology that lacks the necessary data dependencies. The optimal setting at 0.85 indicates that the graph must filter out noise while preserving core structural layout features.

The parameter $\lambda$ controls seed selection for building subgraphs during subproblem solving. It decides how representative an initial vertex must be before the subgraph search begins. SCB reaches 71.3\% accuracy at $\lambda=0.6$, while SB peaks at 26.1\% when $\lambda=0.5$. This difference matches the difficulty of the two datasets. SCB tasks are more regular, so a stricter seed threshold helps limit unnecessary graph expansion. In contrast, a lower $\lambda$ is preferred for the noisier SB dataset because it expands the initial search frontier to prevent valid data paths from being pruned too early. Tuning $\lambda$ allows the model to anchor its reasoning within a locally relevant context for each subtask, ensuring stable performance across spreadsheets with different structures.

\subsection{Case Study} The Figure \ref{case} shows a cross-table revenue calculation case in which the value of the discount node, such as 0.15, can be interpreted in multiple ways. Initially, the explorer and programmer follow an intuitive interpretation and incorrectly treat this value as an absolute discount amount. They therefore construct the computation as $revenue = quantity \times retail~price - discount$. This error is not obvious because the formula is syntactically valid and executable, although its semantic assumption conflicts with the numerical scale of the input. The reflector in the SheetCompass framework identifies this conflict through a state check. Using a floating-point value between 0 and 1 as an absolute monetary deduction is inconsistent with basic business practice. Through the state feedback mechanism, the verification triggers an adaptive retry, prompting the explorer to re-anchor the semantics of the graph node and reinterpret it as a discount rate. the system then corrects the computation and generates the business-consistent formula $revenue = quantity \times retail~price \times (1 - discount)$, which passes verification. this correction demonstrates that verification does more than reject invalid outputs. It provides feedback that reshapes the next round of reasoning. The case highlights how SheetCompass combines topology awareness with a closed-loop cognitive ensemble layer, allowing the system to transition from error detection to logical self-repair while remaining robust and interpretable under complex instructions.

\section{Conclusion}
In this paper, we presented SheetCompass, a novel framework designed to bridge the structural and semantic gap in automated spreadsheet reasoning. By transitioning from flat text sequences to a hierarchical graph, we successfully captured the intrinsic topological hierarchies of spreadsheets. Our multi-agent workflow further emulates human expertise by coordinating specialized explorer, programmer, and reflector roles within a robust execution sandbox. Driven by this structured grounding, these agents collaborate to eliminate errors and maintain reasoning stability. Experimental results on the SCB, SB and SheetRM datasets demonstrate that SheetCompass significantly outperforms baseline methods. Specifically, our approach achieves substantial performance leaps under both soft and hard restrictions by effectively resolving complex data dependencies. This work establishes a strong foundation for future research and provides a scalable, highly reliable solution for the next generation of enterprise-grade spreadsheet automation.

\newpage

\appendix
\section{Experimental Setup}
\subsection{Implementation Details}

The implementation of SheetCompass leverages gpt-5 and gpt-4o-mini as the core reasoning engines within the Cognitive Ensemble, with the temperature maintained at $0.2$ and max tokens set to $4096$ to ensure deterministic and stable output during complex logical synthesis. To realize the mathematical framework of the contextual graph, we employ BGE-M3 as the pre-trained transformer $\Phi$ to map heterogeneous raw features into a $1024$-dimensional latent space, providing a unified foundation for measuring correlations between table and column nodes. Within the semantic alignment, we set the confidence threshold $\alpha$ to $0.85$ to strictly govern the establishment of semantic edges $E_{sem}$, ensuring that implicit logical couplings are only anchored when they surpass rigorous verification. Furthermore, the seeding coefficient $\lambda$ for the collaborative filtering mechanism is configured at $0.6$ as demonstrated in our sensitivity analysis, which optimizes the tradeoff between semantic grounding and context compression to prevent prompt overflow. The entire system is integrated within a sandboxed python environment utilizing openpyxl and excel com interfaces for high-fidelity execution and verification of the generated functional scripts.
\subsection{Datasets}
The evaluation is conducted on three specialized benchmarks designed to assess agent capabilities over complex tabular structures: SheetCopilotBench (SCB), SpreadsheetBench (SB), and SheetRM. The SCB dataset, derived from representative real-world software control tasks, comprises 221 samples characterized by rich cross-sheet dependencies, specifically focusing on advanced spreadsheet manipulations such as the automated configuration of formulas, pivot tables, and visualization charts. The SB dataset introduces higher structural heterogeneity, containing 912 real user instructions compiled from online forums where approximately 35.7\% of the workbooks contain multiple tables within a single sheet and 42.7\% exhibit non-standard relational layouts such as nested or missing headers. Furthermore, following an online judge evaluation pipeline, each instruction in SB is paired with three distinct input-output test cases, totaling 2,729 test cases across the dataset to comprehensively test the model against solution overfitting under irregular layout environments. To evaluate the model resilience against practical operational challenges, the SheetRM dataset introduces 180 multi-category tasks curated from professional spreadsheet software examination banks, which are distinguished by long-horizon multi-step reasoning chains and intentionally ambiguous textual requirements that force the agent to perform iterative task planning and reflection. In this work, a representative subset comprising 50\% of the original samples is extracted from the SheetRM dataset, for which ground truth references are manually constructed to facilitate rigorous verification.
\subsection{Evaluation Metrics}

To provide a granular assessment of the framework performance across heterogeneous evaluation environments, our scoring mechanisms are tailored to the distinct verification paradigms of the baseline benchmarks. For the SheetRM and SCB datasets, which evaluate immediate functional success over discrete software control workflows, we employ Pass@1 and Exec@1 to quantify the percentage of tasks that generate syntactically correct, executable scripts on the first attempt. Conversely, for the complex and irregular layouts of the SB dataset where each high-level instruction is validated against an Online Judge-style suite of multiple test cases, we implement Soft Restriction ($S_{soft}$) and Hard Restriction ($S_{hard}$) as advanced robustness criteria. Let $\mathcal{D}$ denote the total set of instructions in the dataset, $\mathcal{T}_i$ represent the set of input-output test cases for a specific instruction $i \in \mathcal{D}$, and $r_{ij}$ signify the resulting execution status of the $j$-th test case. Soft Restriction measures the foundational logic-capturing capability of the agent by granting partial structural credit proportional to the success rate of individual test cases, formalized as:
\begin{equation}
S_{soft} = \frac{1}{|\mathcal{D}|} \sum_{i=1}^{|\mathcal{D}|} \left( \frac{1}{|\mathcal{T}_i|} \sum_{j=1}^{|\mathcal{T}_i|} \mathbb{1}(r_{ij} = \text{ACC}) \right)
\end{equation}
In contrast, Hard Restriction assesses absolute operational reliability and functional robustness by awarding a positive score for an instruction if and only if all its associated test cases return a flawless execution status, emphasized as:
\begin{equation}
S_{hard} = \frac{1}{|\mathcal{D}|} \sum_{i=1}^{|\mathcal{D}|} \mathbb{1}(r_{ij} = \text{ACC}, \forall j \in \{1, 2, \dots, |\mathcal{T}_i|\})
\end{equation}
where $\mathbb{1}(\cdot)$ serves as the standard indicator function that outputs 1 when its internal condition is satisfied and 0 otherwise. By juxtaposing immediate execution indicators with these multi-case restriction constraints, the evaluation framework effectively isolates simple code memorization from genuine spatial layout reasoning.

\section{Prompt Specifications}

\subsection{Prompt for Semantic Alignment}

\noindent [System] You are a Spreadsheet Knowledge Graph Expert. Your objective is to perform semantic deconstruction and dependency scoring between spreadsheet nodes by providing concise semantic profiles and quantifying relationship confidence. [User] Instruction <task\_description> Cross-table edges <physical\_edges> Nodes <node\_list>. [Reasoning Objectives] (1) Node Profile: Generate a task-aware semantic phrase for each <node\_id> reflecting its functional role within the spreadsheet hierarchy. (2) Relationship Scoring: Identify potential semantic links such as derive  or align. (3) Quantification: For each link, assign a score between 0.0 and 1.0 representing the probability of the relationship's existence considering the provided spreadsheet context. [Output Specification] Return a flattened list containing node\_semantics with their corresponding functional phrases and semantic\_scoring entries, including source node, target node, link type, confidence score, and a brief logical rationale.

\subsection{Prompt for the Explorer Agent}
\noindent [System] You are a Spreadsheet Analyzer. Your objective is to first break the high-level instruction into ordered steps, and then explore the spreadsheet graph by invoking specialized tools. Return JSON only. [User] Instruction <instruction> Current step <step> Seed ids <seeds> Current subgraph <subgraph> History <history> Budget <turn, max\_turns, max\_nodes>. [Reasoning Objectives] (1) Task Decomposition: Deconstruct the goal into atomic reasoning steps. (2) Topological Exploration: Based on the current step and history, select one tool from [search\_nodes, get\_neighborhood, finish\_exploration] to extract relevant subgraphs. (3) Contextual Grounding: Ensure that each exploration move aligns with the specific data dependencies required by the decomposed steps. [Output Specification] Return a JSON object that either defines the reasoning path in a steps key: \{ "steps": ["<step1>", "<step2>"] \}, or executes a graph operation: \{ "tool": "search\_nodes, get\_neighborhood, finish\_exploration", "args": \{\} \}.

\subsection{Prompt for Programmer Agent}
[System] You are an advanced code-generation agent capable of cross-modal spreadsheet reasoning. Your objective is to translate operational instructions and structural tabular contexts into python code through a multi-turn, tool-augmented reasoning loop. [User] Task Specification <steps> Structural Subgraph <shared subgraph> [Reasoning Objectives] (1) Contextual Mapping: Analyze the structural subgraph to align the semantic entities and anchoring coordinates with the required execution logic. (2) Algorithmic Synthesis: Translate the decomposed steps index and target instruction into precise programmatic actions. (3) Structural Preservation: Ensure the synthesized code strictly respects absolute spreadsheet coordinates and maintains topological integrity throughout execution. [Output Specification] Return a single structured JSON object specifying the designated tool identifier and its operational parameters to interact with the environment until task completion.
\subsection{Prompt for Reflector Agent}
[System] You are a unified spreadsheet verification and repair agent. Your objective is to evaluate whether the generated code satisfies the target instructions by analyzing physical runtime execution signals. [User] Task Specification  Runtime Execution Signals <execution state>. [Reasoning Objectives] (1) Empirically Grounded Evaluation: Assess structural modifications by cross-referencing execution states with multi-point workbook telemetry data including before and after statistics and diff summaries to determine semantic alignment. (2) Quantitative Verdict Assignment: Map the evaluated system state to an explicit constraint token representing structural validity such as pass, fail or suspicious.

\section{Expert Knowlege Memory}
\subsection{Domain Knowledge}
The domain knowledge base utilizes a comprehensive three tier architecture consisting of static rules, proactive hints, and evolutionary failure patterns to deliver logical constraints and semantic priors across diverse spreadsheet tasks. Within this framework, the combination of static rules and proactive hints serves as the foundational line of defense to ensure operational safety and logical alignment. Specifically, the system prevents destructive circular reference loops by strictly excluding target destination cells from data fetching and statistical functions. To maintain the integrity of business logic, the system enforces precise behavioral mappings, ensuring that data filtering translates to automated row hiding rather than physical row deletion, and visual highlighting maps to dynamic conditional formatting rather than static cell fills. Furthermore, before any specific tool invocation occurs, the system performs an automated semantic scan to detect high risk business intents, such as complex matrix level extrema calculations, and immediately injects high priority alerts to intercept common technical pitfalls. Complementing these rigid architectural constraints is an adaptive evolutionary failure pattern mechanism that monitors a live catalog of runtime execution errors, including type mismatches, input output faults, spatial anchoring shifts, and resource leaks. When a specific failure pattern or a valuable lesson verified during the reflection phase exceeds a predefined frequency threshold, the system triggers a promotion mechanism that serializes these insights into long term storage. This continuous optimization loop converts real time runtime experiences into universal static constraints.
\subsection{Tool-Use Knowledge}
This framework categorizes the execution media into two distinct tool types consisting of a script control backend based on a high level programming language and a host application backend driven by native spreadsheet software. Both modules are integrated into the system via always on loading or keyword triggered activation mechanisms depending on operational requirements. The script control backend is primarily utilized for low latency and stateless manipulation of raw file structures, focusing on high efficiency incremental data writing and standard style configurations. However, this script tool explicitly lacks live execution time computing capabilities native to the spreadsheet engine, meaning it cannot handle dynamic formula evaluation, automatic column width adjustments, or native pivot table creation. To prevent accidental data corruption or full table overwrites, operations under this backend must avoid assuming that target tables begin at the origin cell, requiring the system to dynamically anchor absolute coordinates based on observable schema metadata before writing data. Conversely, the host application backend is dedicated to managing complex tasks that depend heavily on the internal spreadsheet engine, including the generation of interactive graphical objects and runtime analytical computations. Operating through a standard component object model interface to drive the native spreadsheet application, this tool provides a robust execution environment that can handle advanced user macros and deep formatting dependencies. Because driving a native application consumes significant system memory, this backend enforces a comprehensive lifecycle management workflow that governs environment initialization, workbook context establishment, active cell positioning, exception handling, and explicit process resource reclamation. This careful tracking ensures that heavy background processes are properly terminated.



\section*{GenAI Usage Disclosure}
In accordance with the ACM Policy on the Use of AI, we fully disclose the utilization of Generative AI tools. Codex was employed to assist in writing the baseline evaluation scripts as well as specific implementation portions of our framework code, and ChatGPT was used during the writing stage exclusively to improving grammatical correctness and refining the text flow. The final phrasing, technical accuracy, and scientific arguments remain entirely the responsibility of the authors.

\bibliographystyle{ACM-Reference-Format}
\bibliography{main}


\end{document}